\documentclass[letterpaper, 10 pt, conference, twoside]{support/IEEEconf}

\let\labelindent\relax

\usepackage{times}
\usepackage[pdftex]{graphicx}
\usepackage{subfigure}
\usepackage{amsmath,amssymb,amsopn,amstext,amsfonts}
\usepackage{cancel}
\usepackage[space]{cite}
\usepackage{soul}
\usepackage{cuted}
\usepackage{lipsum,multicol}
\usepackage{stfloats}
\everymath{\displaystyle}
\usepackage{booktabs}
\usepackage{amsthm}
\usepackage{xfrac}
\usepackage{balance}
\usepackage{color}
\usepackage{mathtools}
\usepackage{tabularx}
\usepackage{bm}
\usepackage{diagbox}
\usepackage{float}
\usepackage{epstopdf}
\usepackage{url}
\usepackage{multirow}
\usepackage{tikz}
\usepackage[linkcolor=black,citecolor=black,urlcolor=black,colorlinks=true]{hyperref}
\usepackage{enumitem}
\usepackage[ruled,vlined,linesnumbered]{algorithm2e}

\DeclareMathAlphabet\mathbfcal{OMS}{cmsy}{b}{n}

\soulregister\cite7
\soulregister\citep7
\soulregister\citet7
\soulregister\ref7
\soulregister\it7
\soulregister\pageref7

\usepackage{arydshln}
\graphicspath{{figures/}}
\DeclareGraphicsExtensions{.pdf,.png,.jpg,.eps}

\IEEEoverridecommandlockouts

\title{\LARGE \bf FAST-LIO: A Fast, Robust LiDAR-inertial Odometry Package by Tightly-Coupled Iterated Kalman Filter}
\author{Wei Xu$^{1}$, Fu Zhang$^{1}$
\thanks{This work is funded by a DJI (project no.200009538).}
\thanks{$^{1}$All authors are with Mechatronics and Robotic Systems (MaRS) Laboratory, Department of Mechanical Engineering, University of Hong Kong. {\tt\small \{xuweii, fuzhang\}@hku.hk}}
}

\begin{document}
\maketitle
\begin{abstract}
This paper presents a computationally efficient and robust LiDAR-inertial odometry framework. We fuse LiDAR feature points with IMU data using a tightly-coupled iterated extended Kalman filter to allow robust navigation in fast-motion, noisy or cluttered environments where degeneration occurs. To lower the computation load in the presence of a large number of measurements, we present a new formula to compute the Kalman gain. The new formula has computation load depending on the state dimension instead of the measurement dimension. The proposed method and its implementation are tested in various indoor and outdoor environments. In all tests, our method produces reliable navigation results in real-time: running on a quadrotor onboard computer, it fuses more than 1,200 effective feature points in a scan and completes all iterations of an iEKF step within 25 ms. Our codes are open-sourced on Github\footnote[2]{\url{https://github.com/hku-mars/FAST_LIO}}.
\end{abstract}


\section{Introduction}
Simultaneous localization and mapping (SLAM) is a fundamental prerequisite of mobile robots, such as unmanned aerial vehicles (UAVs). Visual (-inertial) odometry (VO), such as Stereo VO~\cite{sun2018robust} and Monocular VO~\cite{qin2018vins,forster2014svo} are commonly used on mobile robots due to their lightweight and low-cost. Although providing rich RGB information, visual solutions lack direct depth measurements and require much computation resources to reconstruct the 3D environment for trajectory planning. Moreover, they are very sensitive to lighting conditions. Light detection and ranging (LiDAR) sensors could overcome all these difficulties but have been too costly (and bulky) for small-scale mobile robots.

Solid-state LiDARs recently emerge as main trends in LiDAR developments, such as those based on micro-electro-mechanical-system (MEMS) scanning~\cite{wang2020mems} and rotating prisms~\cite{liu2020low}. These LiDARs are very cost-effective (in a cost range similar to global shutter cameras), lightweight (can be carried by a small-scale UAV), and of high performance (producing active and direct 3D measurements of long-range and high-accuracy). These features make such LiDARs viable for UAVs, especially industrial UAVs, which need to acquire accurate 3D maps of the environments (e.g., aerial mapping) or may operate in cluttered environments with severe illumination variations (e.g., post-disaster search and inspection).

Despite the great potentiality, solid-state LiDARs bring new challenges to SLAM: 1) the feature points in LiDAR measurements are usually the geometrical structures (e.g., edges and planes) in the environments. When the UAV is operating in cluttered environments where no strong features are present, the LiDAR-based solution easily degenerates. This problem is more obvious when the LiDAR has a small FoV. 2) Due to the high-resolution along the scanning direction, a LiDAR scan usually contains many feature points (e.g., a few thousand). While these feature points are not adequate to reliably determine the pose in case of degeneration, tightly fusing such a large number of feature points to IMU measurements requires tremendous computational resources that are not affordable by the UAV onboard computer. 3) Since the LiDAR samples point sequentially with a few laser/receiver pairs, laser points in a scan are always sampled at different times, resulting in motion distortion that will significantly degrade a scan registration \cite{besl1992method}. The constant rotations of UAV propellers and motors also introduce significant noises to the IMU measurements.
\begin{figure}[t]
    \begin{center}
        {\includegraphics[width=1.0\columnwidth]{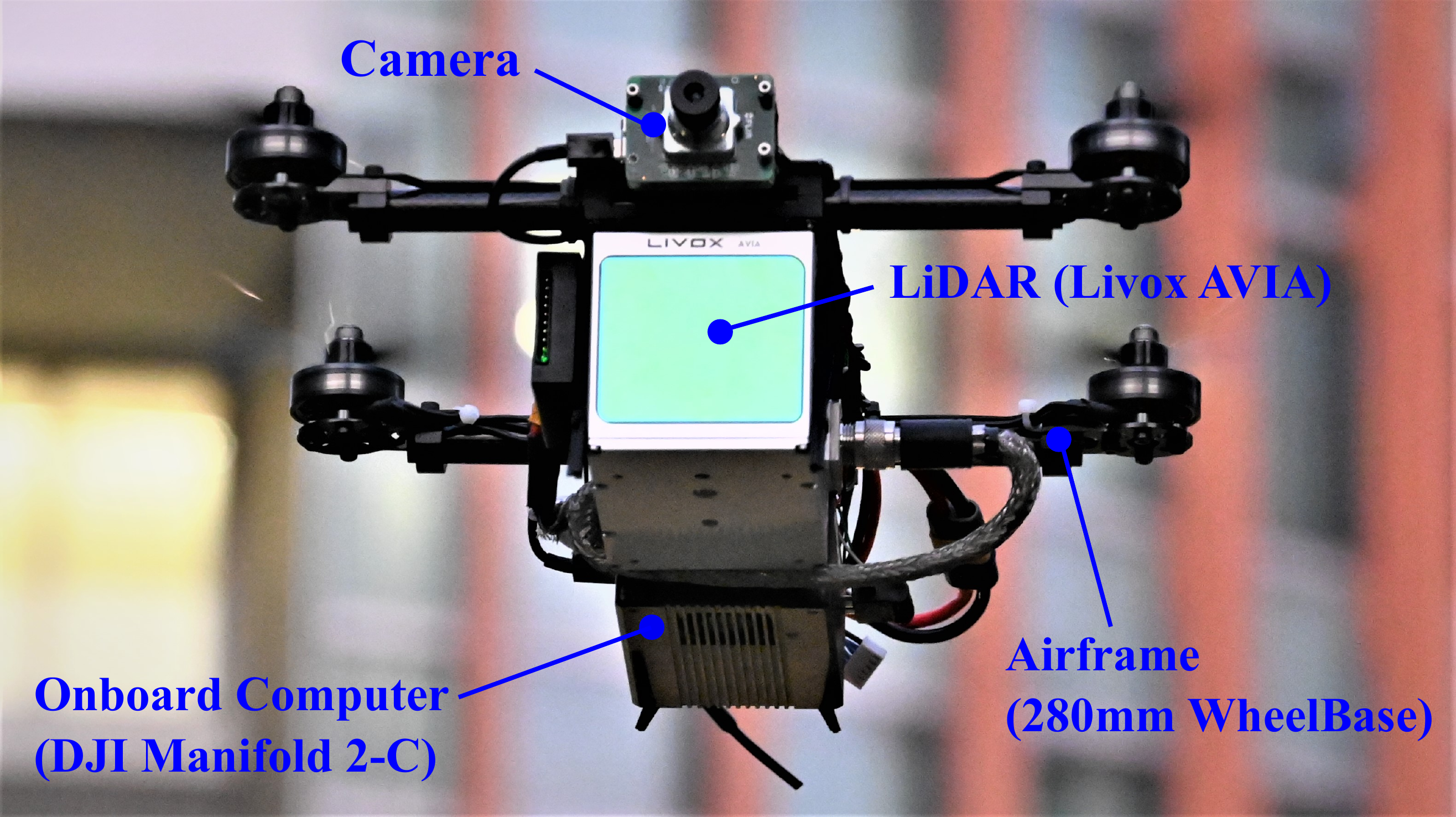}}
    \end{center}
    \caption{\label{fig:uav_fly} {Our LiDAR-inertial navigation system runs on a Livox AVIA LiDAR\protect\footnotemark[3] and a DJI Manifold 2-C onboard computer\protect\footnotemark[4], all on a customized small-scale quadrotor UAV (280 $mm$ wheelbase). The RGB camera is not used in our algorithm, but only for visualization. Video is available at \url{https://youtu.be/iYCY6T79oNU}}}
\end{figure}
\footnotetext[3]{\url{https://www.livoxtech.com/de/avia}}
\footnotetext[4]{\url{https://www.dji.com/cn/manifold-2/specs}}

To make the LiDAR navigation viable for small-scale mobile robots such as UAVs, we propose the FAST-LIO, a computationally efficient and robust LiDAR-inertial odometry package. More specifically, our contributions are as follows: 1) To cope with fast-motion, noisy or cluttered environments where degeneration occurs, we adopt a tightly-coupled iterated Kalman filter to fuse LiDAR feature points with IMU measurements. We propose a formal back-propagation process to compensate for the motion distortion; 2) To lower the computation load caused by a large number of LiDAR feature points, we propose a new formula for computing the Kalman gain and prove its equivalence to the conventional Kalman gain formula. The new formula has a computation complexity depending on the state dimension instead of the measurement dimension. 3) We implement our formulations into a fast and robust LiDAR-inertial odometry software package. The system is able to run on a small-scale quadrotor onboard computer. 4) We conduct experiments in various indoor and outdoor environments and with actual UAV flight tests (Fig.~\ref{fig:uav_fly}) to validate the system's robustness when fast motion or intense vibration noise exists.

The remaining paper is organized as follows: In Section. \ref{sec:related_work}, we discuss relevant research works. We give an overview of the complete system pipeline and the details of each key component in Section. \ref{sec:method}. The experiments are presented in Section. \ref{sec:experiment}, followed by conclusions in Section. \ref{sec:conclusion}.

\section{Related Works}\label{sec:related_work}

Existing works on LiDAR SLAM are extensive. Here we limit our review to the most relevant works: LiDAR-only odometry and mapping, loosely-coupled and tightly-coupled LiDAR-Inertial fusion methods.

\subsection{LiDAR Odometry and Mapping}

Besl {\it {et al.}}~\cite{besl1992method} propose an iterated closest points (ICP) method for scan registration, which builds the basis for LiDAR odometry. ICP performs well for dense 3D scans. However, for sparse point clouds of LiDAR measurements, the exact point matching required by ICP rarely exists. To cope with this problem, Segal {\it et al.}~\cite{segal2009generalized} propose a generalized-ICP based on the point-to-plane distance. Then Zhang {\it et al.}~\cite{zhang2014loam} combine this ICP method with a point-to-edge distance and developed a LiDAR odometry and mapping (LOAM) framework. Thereafter, many variants of LOAM have been developed, such as LeGO-LOAM~\cite{shan2018lego} and LOAM-Livox~\cite{lin2019loam_livox}. While these methods work well for structured environments and LiDARs of large FoV, they are very vulnerable to featureless environments or small FoV LiDARs \cite{lin2019loam_livox}. 

\subsection{Loosely-coupled LiDAR-Inertial Odometry}

IMU measurements are commonly used to mitigate the problem of LiDAR degeneration in featureless environments. Loosely-coupled LiDAR-inertial odometry (LIO) methods typically process the LiDAR and IMU measurements separately and fuse their results later. For example, IMU-aided LOAM~\cite{zhang2014loam} takes the pose integrated from IMU measurements as the initial estimate for LiDAR scan registration. Zhen {\it et al.}~\cite{zhen2017robust} fuse the IMU measurements and the Gaussian Particle Filter output of LiDAR measurements using the error-state EKF. Balazadegan {\it et al}~\cite{balazadegan2016visual} add the IMU-gravity model to estimate the 6-DOF ego-motion to aid the LiDAR scan registration. Zuo {\it et al.}~\cite{zuo2019lic} use a Multi-State Constraint Kalman Filter (MSCKF) to fuse the scan registration results with IMU and visual measurements. A common procedure of the loosely-coupled approach is obtaining a pose measurement by registering a new scan and then fusing the pose measurement with IMU measurements. The separation between scan registration and data fusion reduces the computation load. However, it ignores the correlation between the system's other states (e.g., velocity) and the pose of the new scan. Moreover, in the case of featureless environments, the scan registration could degenerate in certain directions and causes unreliable fusion in later stages.

\subsection{Tightly-coupled LiDAR-Inertial Odometry}

Unlike the loosely-coupled methods, tightly-coupled LiDAR-inertial odometry methods typically fuse the raw feature points (instead of scan registration results) of LiDAR with IMU data. There are two main approaches to tightly-coupled LIO: optimization-based and filter-based. Geneva {\it et al.}~\cite{geneva2018lips} use a graph optimization with IMU pre-integration constrains~\cite{forster2016manifold} and plane constraints~\cite{hsiao2018dense} from LiDAR feature points. Recently, Ye {\it et al.}~\cite{ye2019tightly} propose the LIOM package which uses a similar graph optimization but is based on edge and plane features. For filter-based methods, Bry~\cite{bry2012state} uses a Gaussian Particle Filter (GPF) to fuse the data of IMU and a planar 2D LiDAR. This method has also been used in the Boston Dynamics Atlas humanoid robot. Since the computation complexity of particle filter grows quickly with the number of feature points and the system dimension, Kalman filter and its variants are usually more preferred, such as extended Kalman filter~\cite{hesch2010laser}, unscented Kalman filter~\cite{cheng2015practical}, and iterated Kalman filter \cite{qin2019lins}.

Our method falls into the tightly-coupled approach. We adopt an iterated extended Kalman filter similar to \cite{qin2019lins} to mitigate linearization errors. Kalman filter (and its variants) has a time complexity $\mathcal{O}\left(m^2 \right)$ where $m$ is the measurement dimension~\cite{raitoharju2019computational}, this may lead to remarkably high computation load when dealing with a large number of LiDAR measurements. Naive down-sampling would reduce the number of measurements but at the cost of information loss. \cite{qin2019lins} reduces the number of measurements by extracting and fitting ground planes similar to~\cite{shan2018lego}. This, however, does not apply to aerial applications where the ground plane may not always present.
	
\section{Methodology}\label{sec:method}
\subsection{Framework Overview}
\begin{figure*}[t]
    \begin{center}
        {\includegraphics[width=2.05\columnwidth]{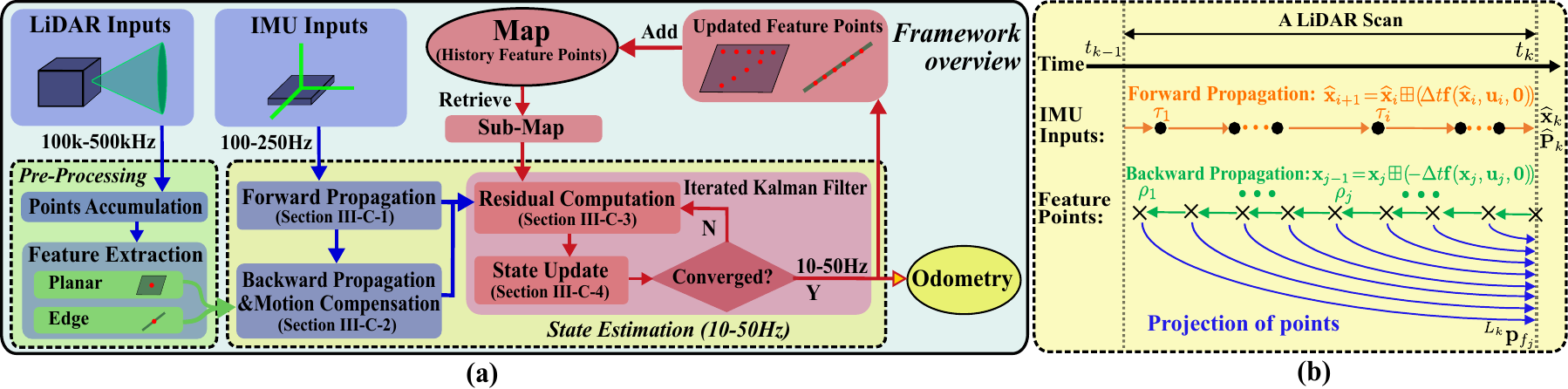}}
    \end{center}
    \vspace{-0.4cm}
    \caption{\label{fig:workflow}{System overview of FAST-LIO. (a): the overall pipeline; (b): the forward and backward propagation.}}
\end{figure*}

{This paper will use the notations shown in Table I.} The overview of our workflow is shown in Fig.~\ref{fig:workflow}. The LiDAR inputs are fed into the feature extraction module to obtain planar and edge features. Then the extracted features and IMU measurements are fed into {our state estimation module for state estimation at $10Hz-50Hz$. The estimated pose then registers the feature points to the global frame and merges them with the feature points map built so far. The updated map is finally used to register further new points in the next step.}

\begin{table}[htbp]
\centering
\caption{{Some Important Notations}}
\label{tab::c_runtime}
\begin{tabularx}{0.49\textwidth}{lX}
\toprule
Symbols & Meaning \\ \midrule
$t_k$ & The scan-end time of the $k$-th LiDAR scan.\\
$\tau_i$ & The $i$-th IMU sample time in a LiDAR scan.\\
$\rho_j$ & The $j$-th feature point's sample time in a LiDAR scan.\\
$I_i, I_j, I_k$ & The IMU body frame at the time $\tau_i$, $\rho_j$ and $t_k$.\\
$L_j, L_k$ & The LiDAR body frame at the time $\rho_j$ and $t_k$.\\
$\mathbf x, \widehat{\mathbf x}, \bar{\mathbf x}$ & The ground-true, propagated, and updated value of $\mathbf x$. \\
$\widetilde{\mathbf x}$ & The error between ground-true $\mathbf x$ and its estimation $\bar{\mathbf x}$.\\
$\widehat{\mathbf x}^\kappa$ & The $\kappa$-th update of $\mathbf x$ in the iterated Kalman filter.\\
${\mathbf x}_{i}, \mathbf x_j, \mathbf x_k$ & The vector (e.g.,state) $\mathbf x$ at time $\tau_i, \rho_j$ and $t_k$.\\
$\check{\mathbf x}_j$ & Estimate of $\mathbf x_j$ relative to $\mathbf x_k$ in the back propagation. \\
\bottomrule
\end{tabularx}
\end{table}

\subsection{System Description}\label{sec:state_equation}
\subsubsection{$\boxplus$ / $\boxminus$ operator}
\
\par
{Let $\mathcal{M}$ be the manifold of dimension $n$ in consideration (e.g., $\mathcal{M} = SO(3)$). Since manifolds are locally homeomorphic to $\mathbb{R}^n$, we can establish a bijective mapping from a local neighborhood on $\mathcal{M}$ to its tangent space $\mathbb{R}^n$ via two encapsulation operators $\boxplus$ and $\boxminus$}~\cite{hertzberg2013integrating}:
\begin{equation}\notag
\begin{aligned}
    &\ \ \ \ \ \ \ \ \ \ \ \boxplus\!:\! \mathcal{M}\! \times\! \mathbb{R}^n \!\!\rightarrow\!\! \mathcal{M}; \ \ \ \ \ \ \boxminus\!:\!\mathcal{M} \!\times\! \mathcal{M} \!\rightarrow \!\mathbb{R}^n\\
    &\mathcal{M}\!\!=\!SO(3): \mathbf{R}\!\boxplus\!\mathbf{r}\!=\!\mathbf{R} \rm{Exp}(\mathbf{r});\ \ \ \ \mathbf{R}_1\!\!\boxminus\!\mathbf{R}_2\!=\!\rm{Log} (\mathbf{R}_2^{T} \mathbf{R}_1\!) \\
    &\mathcal{M}\!\!=\!\mathbb R^n: \ \ \ \ \ \mathbf{a}\!\boxplus\!\mathbf{b}\!=\mathbf{a} \!+\! \mathbf b;\ \ \ \ \ \ \ \ \  \mathbf{a}\boxminus\mathbf{b}=\mathbf{a}\!-\!\mathbf{b}
\end{aligned}
\end{equation}
{where $\textstyle\text{Exp}\left(\mathbf r\right)\!=\!\mathbf I\!+\!\frac{\mathbf r}{\|\mathbf r\|}\sin\left(\|\mathbf r\|\right)\!+\!\frac{\mathbf r^2}{\|\mathbf r\|^2}\left(1\!-\!\cos\left(\|\mathbf r\|\right)\right)$ is the exponential map\cite{hertzberg2013integrating} and $\text{Log}(\cdot)$ is its inverse map. For a compound manifold $\mathcal M = SO(3) \times \mathbb{R}^n$ we have}:

$$\begin{bmatrix}\mathbf R \\ \mathbf a \end{bmatrix} \boxplus \begin{bmatrix}\mathbf r \\ \mathbf b \end{bmatrix} = \begin{bmatrix} \mathbf R\boxplus \mathbf r\\ \mathbf a + \mathbf b \end{bmatrix}; \ \ 
\begin{bmatrix}\mathbf R_1 \\ \mathbf a \end{bmatrix} \boxminus \begin{bmatrix}\mathbf R_2 \\ \mathbf b \end{bmatrix} = \begin{bmatrix} \mathbf R_1\boxminus \mathbf R_2\\ \mathbf a - \mathbf b \end{bmatrix} $$

{From the above definition, it is easy to verify that} 
$$(\mathbf x \! \boxplus \! \mathbf u ) \! \boxminus \! \mathbf x \! = \! \mathbf u; \ 
\mathbf x \! \boxplus \! (\mathbf y \! \boxminus \! \mathbf x) \!= \! \mathbf y; \ \forall \mathbf x, \mathbf y \in \mathcal{M}, \ \forall \mathbf u \in \mathbb{R}^n. $$


\subsubsection{Continuous model}\label{sec:state_equation_cont}
\
\par
Assuming an IMU is rigidly attached to the LiDAR with a known extrinsic $^{I}\mathbf T_{L}=\left(^I\mathbf R_{L},\;^I\mathbf p_L\right)$. Taking the IMU frame (denoted as $I$) as the body frame of reference leads to a kinematic model:
\begin{equation}~\label{e:kine_model_s}
\begin{aligned}
^G\dot{\mathbf p}_I&={}^G\mathbf v_I,\ ^G\dot{\mathbf v}_I={}^G{\mathbf R}_I \left( \mathbf a_m - \mathbf b_{\mathbf a} - \mathbf n_{\mathbf a} \right) + {}^G\mathbf g,\ ^G\dot{\mathbf g} = \mathbf 0\\
^G\dot{\mathbf R}_I&={}^G{\mathbf R}_I \lfloor \bm \omega_m -  \mathbf b_{\bm \omega} - \mathbf n_{\bm \omega} \rfloor_\wedge,\ \dot{\mathbf b}_{\bm \omega}=\mathbf n_{\mathbf b\bm \omega}, \ \dot{\mathbf b}_{\mathbf a}=\mathbf n_{\mathbf b\mathbf a}
\end{aligned}
\end{equation}
where $^G\mathbf p_I$, $^G{\mathbf R}_I$ are the position and attitude of IMU in the global frame (i.e., the first IMU frame, denoted as $G$), $^G\mathbf g$ is the unknown gravity vector in the global frame, $\mathbf a_m$ and $\bm \omega_m$ are IMU measurements, $\mathbf n_{\mathbf a}$ and $\mathbf n_{\bm \omega}$ are the white noise of IMU measurements, $\mathbf b_{\mathbf a}$ and $\mathbf b_{\boldsymbol{\omega}}$ are the IMU bias modelled as the random walk process with Gaussian noises $\mathbf n_{\mathbf b \mathbf a}$ and $\mathbf n_{\mathbf b\bm \omega}$, and the notation $\lfloor \mathbf a \rfloor_{\wedge}$ denotes the skew-symmetric matrix of vector $\mathbf a \in \mathbb{R}^3$ {that maps the cross product operation}.

\subsubsection{Discrete model}
\
\par
Based on the $\boxplus$ operation defined above, we can discretize the continuous model in (\ref{e:kine_model_s}) at the IMU sampling period $\Delta t$ using a zero-order holder. The resultant discrete model is
\begin{equation}\label{e:kine_model_discrete}
    \begin{aligned}
    \mathbf x_{i+1} &= \mathbf x_{i} \boxplus \left( \Delta t \mathbf f(\mathbf x_i, \mathbf u_i, \mathbf w_i) \right)
    \end{aligned}
\end{equation}
where $i$ is the index of IMU measurements, the function $\mathbf f$, state $\mathbf x$, input $\mathbf u$ and noise $\mathbf w$ are defined as below:
\begin{equation}
\begin{split}~\label{e:states_k}
    \mathcal{M} &= SO(3) \times \mathbb{R}^{15}, \ \text{dim}(\mathcal{M}) = 18 \\
    \mathbf x &\doteq \begin{bmatrix}^G{\mathbf R}_I^T&^G\mathbf p_I^T&^G\mathbf v_I^T&\mathbf b_{\bm \omega}^T&\mathbf b_{\mathbf a}^T&^G\mathbf g^T\end{bmatrix}^T \in \mathcal{M} \\
    \mathbf u &\doteq \begin{bmatrix} {\bm \omega}^T_m & {\mathbf a}^T_m\end{bmatrix}^T,\
    \mathbf w \doteq \begin{bmatrix}\mathbf n_{\bm \omega}^T&\mathbf n_{\mathbf a}^T&\mathbf n_{\mathbf b\bm \omega}^T&\mathbf n_{\mathbf b\mathbf a}^T\end{bmatrix}^T \\
    \mathbf f&\left( \mathbf x_{i}, \mathbf u_i, \mathbf w_i\right) = \begin{bmatrix} \bm \omega_{m_i} - \mathbf b_{\bm \omega_i} - \mathbf n_{\bm \omega_i} \\
    {}^G\mathbf v_{I_i} \\
    {}^G{\mathbf R}_{I_i}\left(\mathbf a_{m_i} - \mathbf b_{\mathbf a_i}- \mathbf n_{\mathbf a_i}\right) + {}^G\mathbf g_i  \\
    \mathbf n_{\mathbf b \bm \omega_i} \\
    \mathbf n_{\mathbf b \mathbf a_i} \\
    \mathbf 0_{3\times1} \end{bmatrix}
\end{split}
\end{equation}

\subsubsection{Preprocessing of LiDAR measurements}\label{secMeas}
\
\par
{LiDAR measurements are point coordinates in its local body frame.} Since the {raw LiDAR points} are sampled at a very high rate (e.g., 200kHz), it is usually not possible to process each new point once being received. A more practical approach is to accumulate these points for a certain time and process them all at once. {In FAST-LIO, the minimum accumulation interval is set to 20 $ms$, leading to up to 50 $Hz$ full state estimation (i.e., odometry output) and map update as shown in Fig.~\ref{fig:workflow} (a).} Such an accumulated set of points is called a scan, and the time for processing it is denoted as $t_k$ (see Fig. \ref{fig:workflow} (b)). {From the raw points, we extract planar points with high local smoothness\cite{zhang2014loam} and edge points with low local smoothness as in\cite{lin2019loam_livox}. Assume the number of feature points is $m$, each is sampled at time $\rho_j \in (t_{k-1}, t_{k}]$ and is denoted as ${}^{L_j}\mathbf p_{f_j}$, where $L_j$ is the LiDAR local frame at the time $\rho_j$}. During a LiDAR scan, there are also multiple IMU measurements, each sampled at time $\tau_i \in [t_{k-1}, t_{k}]$ with the respective state $\mathbf x_{i}$ as in (\ref{e:kine_model_discrete}). {Notice that the last LiDAR feature point is the end of a scan, i.e., $\rho_m = t_k$, while the IMU measurements may not necessarily be aligned with the start or end of the scan. }

\subsection{State Estimation}\label{sec:stat_est}
To estimate the states in the state formulation (\ref{e:kine_model_discrete}), we use an iterated extended Kalman filter. Moreover, we characterize the estimation covariance in the tangent space of the state estimate as in \cite{hertzberg2013integrating, xu2020robots}. Assume the optimal state estimate of the last LiDAR scan at $t_{k-1}$ is $\bar{\mathbf {x}}_{k-1}$ with covariance matrix $\bar{\mathbf P}_{k-1}$. Then $\bar{\mathbf P}_{k-1}$ represents the covariance of the random error state vector defined below:
\begin{equation}~\label{e:error_state_def}\notag
    \widetilde{\mathbf x}_{k-1} \!\doteq\! {\mathbf x}_{k-1} \boxminus \bar{\mathbf x}_{k-1} \!=\! \begin{bmatrix} \delta \bm \theta^T \!\!&\!\! {}^G\widetilde{\mathbf p}_I^T \!&\! {}^G\widetilde{\mathbf v}^T_I&\widetilde{\mathbf b}_{\bm \omega}^T \!&\!  \widetilde{\mathbf b}_{\mathbf a}^T \!&\! {}^G\widetilde{\mathbf g}^T\end{bmatrix}^T
\end{equation}
where $\delta \bm \theta=\text{Log}({}^G\bar{\mathbf R}_I^T\mathbf {}^G \mathbf R_I )$ is the attitude error and the rests are standard additive errors (i.e., the error in the estimate $\bar{\mathbf x} $
of a quantity $\mathbf x$ is $\widetilde{\mathbf x} = \mathbf x - \bar{\mathbf x}$). Intuitively, the attitude error $\delta \bm \theta $ describes the (small) deviation between the true and the estimated attitude. The main advantage of this error definition is that it allows us to represent the attitude uncertainty by the $3\times3$ covariance matrix $ \mathbb{E} \left\{\delta\bm\theta\delta\bm\theta^T\right\}$. Since the attitude has 3 degree of freedom (DOF), this is a minimal representation.

\subsubsection{Forward Propagation}\label{sec:forward_prog}
\
\par
The forward propagation is performed once receiving an IMU input (see Fig.~\ref{fig:workflow}). More specifically, the state is propagated following (\ref{e:kine_model_discrete}) by setting the process noise $\mathbf w_i$ to zero:
\begin{equation}\label{e:state_prop}
    \begin{aligned}
    \widehat{ \mathbf  x}_{i+1} &= \widehat{ \mathbf  x}_{i} \boxplus \left( \Delta t \mathbf f (\widehat{\mathbf x}_i, \mathbf u_i, \mathbf 0) \right); \ \widehat{\mathbf x}_{0} = \bar{\mathbf x}_{k-1}. 
    \end{aligned}
\end{equation}
where $\Delta t = \tau_{i+1} - \tau_{i}$. To propagate the covariance, we use the error state dynamic model obtained below:
\begin{equation}
\begin{split}~\label{e:error_state_model}
\widetilde{\mathbf x}_{i+1} \!&=\! {\mathbf x}_{i+1} \boxminus \widehat{\mathbf x}_{i+1} \\
    &=\! \left( \mathbf x_{i}\! \boxplus\! \Delta t \mathbf f\left( {\mathbf x}_{i}, \mathbf u_i, \mathbf w_i\right) \right) \boxminus \left(\widehat{\mathbf x}_{i}\! \boxplus\! \Delta t \mathbf f\left( \widehat{\mathbf x}_{i}, \mathbf u_i, \mathbf 0 \right) \right) \\
    &\overset{(\ref{e:F_def})}{\simeq} \mathbf F_{\widetilde{\mathbf x}} \widetilde{\mathbf x}_i + \mathbf F_{\mathbf w} \mathbf w_i.
\end{split}
\end{equation}

The matrix $\mathbf F_{\widetilde{\mathbf x}}$ and $\mathbf F_{\mathbf w}$ in (\ref{e:error_state_model}) is computed following the Appendix.~\ref{Appendix_F_comp}. The result is shown in (\ref{e:F_compute}), {where $\widehat{\bm \omega}_i={\bm \omega}_{m_i} - \widehat{\mathbf b}_{\bm \omega_i}$, $\widehat{\mathbf a}_i={\mathbf a}_{m_i} - \widehat{\mathbf b}_{\mathbf a_i}$ and $\mathbf{A}\left(\mathbf{u}\right)^{-1}$ follows the same definition in \cite{bullo1995proportional} as below:}
\begin{equation}
\begin{split}~\label{e:F_def_parts}
\mathbf{A}\left(\mathbf{u}\right)^{-1}&=\textstyle \mathbf{I}-\frac{1}{2}\lfloor\mathbf{u}\rfloor_{\wedge}+\left(1-\bm\alpha\left(\Vert\mathbf{u}\Vert\right)\right)\frac{\lfloor\mathbf{u}\rfloor_{\wedge}^2}{\Vert\mathbf{u}\Vert^2} \\\alpha\left(\rm{m}\right)&=\textstyle\frac{\rm{m}}{2}{\rm cot}\left(\frac{\rm{m}}{2}\right)=\frac{\rm{m}}{2}\frac{{\rm cos}\left(\rm{m}/2\right)}{{\rm sin}\left(\rm{m}/2\right)}
\end{split}
\end{equation}
\begin{figure*}
\begin{align}~\label{e:F_compute}
\mathbf F_{\widetilde{\mathbf x}}\!\!=\!\!\begin{bmatrix} \text{Exp}\left(-\widehat{\bm \omega}_i\Delta t\right)\!\! & \mathbf 0 & \mathbf 0 &\!\!\!\!-\mathbf A\!\!\left(\widehat{\bm \omega}_i\Delta t\right)^{T}\!\!\Delta t\!\!\!\!& \mathbf 0 & \mathbf 0 \\
\mathbf 0 & \mathbf I &\!\!\!\!\mathbf I\Delta t & \mathbf 0 & \mathbf 0 & \mathbf 0 \\
{-}^G\widehat{\mathbf R}_{I_i}\lfloor\widehat{\mathbf a}_{i}\rfloor_{\wedge}\Delta t & \mathbf 0 & \mathbf I & \mathbf 0 & \!\!-{}^G\widehat{\mathbf R}_{I_i}\Delta t\!\!& \!\!\mathbf I\Delta t \\
\mathbf 0 & \mathbf 0 & \mathbf 0 & \mathbf I & \mathbf 0 & \mathbf 0 \\
\mathbf 0 & \mathbf 0 & \mathbf 0 & \mathbf 0 & \mathbf I & \mathbf 0 \\
\mathbf 0 & \mathbf 0 & \mathbf 0 & \mathbf 0 & \mathbf 0 & \mathbf I 
    \end{bmatrix}\!,\ 
\mathbf F_{\mathbf w}\!\!=\!\! \begin{bmatrix} -\mathbf A\left(\widehat{\bm \omega}_i \Delta t\right)^{T}\Delta t & \mathbf 0 & \mathbf 0 & \mathbf 0 \\
\mathbf 0 & \mathbf 0 & \mathbf 0 & \mathbf 0 \\
\mathbf 0 & -{}^G\widehat{\mathbf R}_{I_i}\Delta t & \mathbf 0 & \mathbf 0 \\
\mathbf 0 & \mathbf 0 & \mathbf I\Delta t & \mathbf 0 \\
\mathbf 0 & \mathbf 0 & \mathbf 0 & \mathbf I\Delta t \\
\mathbf 0 & \mathbf 0 & \mathbf 0 & \mathbf 0 \end{bmatrix}
\end{align}
\hrulefill
\end{figure*}

Denoting the covariance of white noises $\mathbf w$ as $\mathbf Q$, then the propagated covariance $\widehat{\mathbf P}_i$ can be computed iteratively following the below equation.
\begin{equation}
\begin{split}~\label{e:noise_cov}
\widehat{\mathbf P}_{i+1} = \mathbf F_{\widetilde{\mathbf x}}\widehat{\mathbf P}_{i}\mathbf F_{\widetilde{\mathbf x}}^T + \mathbf F_{\mathbf w} \mathbf Q \mathbf F_{\mathbf w}^T ; \ \widehat{\mathbf P}_{0} = \bar{\mathbf P}_{k-1}.
\end{split}
\end{equation}

The propagation continues until reaching the end time of a new scan at $t_{k}$ where the propagated state and covariance are denoted as $\widehat{\mathbf{x}}_{k}, \widehat{\mathbf P}_{k}$.  Then $\widehat{\mathbf P}_{k}$ represents the covariance of the error between the ground-truth state $\mathbf x_k$ and the state propagation $\widehat{\mathbf{x}}_{k}$ (i.e., $\mathbf x_k \boxminus \widehat{\mathbf{x}}_{k} $). 
\subsubsection{Backward Propagation and Motion Compensation}
\
\par
When the points accumulation time interval is reached at time $t_{k}$, the new scan of feature points should be fused with the propagated state $\widehat{\mathbf{x}}_{k}$ and covariance $\widehat{\mathbf P}_{k}$ to produce an optimal state update. However, although the new scan is at time $t_{k}$, the feature points are measured at their respective sampling time $\rho_j \leq t_{k}$ (see Section. \ref{secMeas} and Fig. \ref{fig:workflow} (b)), causing a mismatch in the body frame of reference.

To compensate the relative motion (i.e., motion distortion) between time $\rho_j$ and time $t_{k}$, we propagate (\ref{e:kine_model_discrete}) backward as $\check{\mathbf x}_{j-1} = \check{\mathbf x}_{j} \boxplus \left( -\Delta t \mathbf f(\check{\mathbf x}_j, \mathbf u_j, \mathbf 0) \right)$, starting from zero pose and rests states (e.g., velocity and bias) from $\widehat{\mathbf{x}}_{k}$. The backward propagation is performed at the frequency of feature point, which is usually much higher than the IMU rate. {For all the feature points sampled between two IMU measurements, we use the left IMU measurement as the input in the back propagation. Furthermore, noticing that the last three block elements (corresponding to the gyro bias, accelerometer bias, and extrinsic) of $\mathbf f(\mathbf x_j, \mathbf u_j, \mathbf 0)$ (see (\ref{e:states_k})) are zeros, the back propagation can be reduced to:}

\begin{equation}~\label{e:back_prop}
\hspace{-0.5cm}
\begin{aligned}
\!^{I_k} \check{\mathbf p}_{I_{j\!-\!1}}\!&\!= {}^{I_k} \check{\mathbf p}_{I_{j}}\! - \! {}^{I_k} \check{\mathbf v}_{I_j} \Delta t, \quad \quad \quad  \quad s.f. \ {}^{I_k} \check{\mathbf p}_{I_{m}} = \mathbf 0; \\
\!^{I_k}\check{\mathbf v}_{I_{j\!-\!1}}\!&\!= {}^{I_k} \check{\mathbf v}_{I_{j}}-{}^{I_k}  \check{\mathbf R}_{I_j} ( \mathbf a_{m_{i-1}} - \widehat{\mathbf b}_{\mathbf a_k} )\Delta t - {}^{I_k} \widehat{\mathbf g}_k \Delta t, \\
& \quad \quad s.f. \ {}^{I_k} \check{\mathbf v}_{I_{m}} = {}^{G} \widehat{\mathbf R}_{I_{k}}^T {}^{G} \widehat{\mathbf v}_{I_{k}},  {}^{I_k} \widehat{\mathbf g}_k = ^{G} \! \! \widehat{\mathbf R}_{I_{k}}^T  {}^{G} \widehat{\mathbf g}_k; \\
^{I_k} \check{\mathbf R}_{I_{j-1}}\!&\!= {}^{I_k} \check{\mathbf R}_{I_j} \text{Exp} ( (\widehat{\mathbf b}_{\bm \omega_k}\!\! - \!\bm \omega_{m_{i-1}}) \Delta t ), \ s.f. \   {}^{I_k} {\mathbf R}_{I_m} \! = \! \mathbf I.
\end{aligned}
\end{equation}
{where $\rho_{j-1} \in [\tau_{i-1}, \tau_{i})$, $\Delta t = \rho_j - \rho_{j-1}$, and {\it s.f.} means ``starting from".}

The backward propagation will produce a relative pose between time $\rho_j$ and the scan-end time $t_k$: ${}^{I_k}\check{\mathbf T}_{ I_j}=\left({}^{I_k}\check{\mathbf R}_{I_j},{}^{I_k}\check{\mathbf p}_{I_j}\right)$. This relative pose enables us to project the local measurement ${}^{L_j}{\mathbf p}_{f_j}$ to scan-end measurement ${}^{L_k} {\mathbf p}_{f_j}$ as follows (see Fig. \ref{fig:workflow}):
\begin{equation}
\begin{split}~\label{e:projection}
{}^{L_k}{\mathbf p}_{f_j}={}^{I}{\mathbf T}_{L}^{-1} {}^{I_k}\check{\mathbf T}_{I_j} {}^{I}{\mathbf T}_{L} {}^{L_j}{\mathbf p}_{f_j}.
\end{split}
\end{equation}
where ${}^{I}{\mathbf T}_{L}$ is the known extrinsic (see Section.~\ref{sec:state_equation_cont}). Then the projected point ${}^{L_k}{\mathbf p}_{f_j}$ is used to construct a residual in the following section.
\subsubsection{Residual computation}\label{sec:residual}
\
\par
With the motion compensation in (\ref{e:projection}), we can view the scan of feature points $\{ {}^{L_k} \mathbf{p}_{f_j} \}$ all sampled at the same time $t_k$ and use it to construct the residual. Assume the current iteration of the iterated Kalman filter is $\kappa$, and the corresponding state estimate is $\widehat{\mathbf x}_k^\kappa$. When $\kappa = 0$, $\widehat{\mathbf x}_k^\kappa = \widehat{\mathbf x}_k$, the predicted state from the propagation in (\ref{e:state_prop}). Then, the feature points $\{ ^{L_k} \mathbf{p}_{f_j} \}$ {can be transformed} to the global frame as below:
\begin{equation}
\begin{split}~\label{e:feat_global}
    {}^G \widehat{\mathbf p}_{f_j}^\kappa = {}^G\widehat{\mathbf T}_{I_k}^\kappa {}^I\mathbf T_L  {}^{L_k} \mathbf p_{f_j} ; \ j= 1, \cdots, m.
\end{split}
\end{equation}

{For each LiDAR feature point, the closest plane or edge defined by its nearby feature points in the map is assumed to be where the point truly belongs to. That is, the residual is defined as the distance between the feature point's estimated global frame coordinate $^G \widehat{\mathbf p}_{f_j}^\kappa$ and the nearest plane (or edge) in the map.} Denoting $\mathbf u_j$ the normal vector (or edge orientation) of the corresponding plane (or edge), on which lying a point ${}^G\mathbf q_j$, then the residual $\mathbf z_j^{\kappa}$ is computed as:
\begin{equation}
\begin{split}~\label{e:res_computation}
    \mathbf z_j^\kappa = \mathbf G_j \left( {}^G \widehat{\mathbf p}_{f_j}^\kappa - {}^G\mathbf q_j \right)
\end{split}
\end{equation}
where $\mathbf G_j =\mathbf u_j^T$ for planar features and $\mathbf G_j =\lfloor\mathbf u_j\rfloor_{\wedge}$ for edge features. {The computation of the $\mathbf u_j$ and the search of nearby points in the map, which define the corresponding plane or edge, is achieved by building a KD-tree of the points in the most recent map\cite{lin2019loam_livox}. Moreover, we only consider residuals whose norm is below certain threshold (e.g., $0.5m$). Residuals exceeding this threshold are either outliers or newly observed points. }

\subsubsection{Iterated state update}\label{sec:iter-update}
\
\par
To fuse the residual $\mathbf z_j^\kappa $ computed in (\ref{e:res_computation}) with the state prediction $\widehat{\mathbf{x}}_{k}$ and covariance $\widehat{\mathbf P}_{k}$ propagated from the IMU data, we need to linearize the measurement model that relates the residual $\mathbf z_j^\kappa $ to the ground-truth state $\mathbf x_k$ and {measurement} noise. The measurement noise originates from the LiDAR ranging and beam-directing noise ${}^{L_j} \mathbf n_{f_j}$ when measuring the point ${}^{L_j} \mathbf p_{f_j}$. Removing this noise from the point measurement ${}^{L_j} \mathbf p_{f_j}$ leads to the true point location
\begin{equation}\label{e:gt_meas}
    {}^{L_j} \mathbf p_{f_j}^{\text{gt}} = {}^{L_j} \mathbf p_{f_j} - {}^{L_j} \mathbf n_{f_j}.
\end{equation}
This true point, after projecting to the frame $L_k$ via (\ref{e:projection}) and then to the global frame with the ground-truth state $\mathbf x_k$ (i.e, pose), should lie exactly on the plane (or edge) in the map. That is, plugging (\ref{e:gt_meas}) into (\ref{e:projection}), then into (\ref{e:feat_global}), and further into (\ref{e:res_computation}) should result in zero. i.e.,
\vspace{-0.1cm}
\begin{equation}\label{e:meas_func}\notag
\begin{split}
    \mathbf 0 \! = \! \mathbf h_j \! \left(  \mathbf x_k , \!{}^{L_j}{\mathbf n}_{f_j}\! \right) \!\! = \! \mathbf G_j \!\! \left( \! {}^G {\mathbf T}_{I_k} \! {}^{I_k} \check{\mathbf T}_{I_j} \! {}^{I}{\mathbf T}_{L} \!\! \left( {}^{L_j}{\mathbf p}_{f_j} \! - \! {}^{L_j}{\mathbf n}_{f_j} \! \right) \! \!  - \!\! {}^G\mathbf q_j \!\right)
\end{split}
\end{equation}

Approximating the above equation by its first order approximation made at $\widehat{\mathbf x}_k^\kappa$ leads to
\vspace{-0.1cm}
\begin{equation}\label{e:meas_jocob}
\begin{split}
    \mathbf 0 &= \mathbf h_j \left( \mathbf x_k, {}^{L_j}{\mathbf n}_{f_j} \right) \simeq \mathbf h_j \left( \widehat{\mathbf x}_k^\kappa, \mathbf 0 \right) + \mathbf H_j^\kappa  \widetilde{\mathbf x}_{k}^\kappa + \mathbf v_j\\&=  \mathbf z_j^\kappa + \mathbf H_j^\kappa  \widetilde{\mathbf x}_{k}^\kappa + \mathbf v_j
\end{split}
\end{equation}
where $\widetilde{\mathbf x}_{k}^\kappa=\mathbf x_k \boxminus \widehat{\mathbf x}_k^\kappa$ (or equivalently $\mathbf x_k = \widehat{\mathbf x}_k^\kappa \boxplus \widetilde{\mathbf x}_{k}^\kappa$), $\mathbf H_j^\kappa$ is the Jacobin matrix of {$\mathbf h_j\! \left(  \widehat{\mathbf x}_k^\kappa \boxplus \widetilde{\mathbf x}_{k}^\kappa , {}^{L_j}{\mathbf n}_{f_j} \right)$} with respect to $\widetilde{\mathbf x}_{k}^\kappa$, evaluated at zero, and $\mathbf v_j \in \mathcal{N}(\mathbf 0, \mathbf R_j)$ comes from the raw measurement noise ${}^{L_j}{\mathbf n}_{f_j}$. 

Notice that the prior distribution of $\mathbf x_k$ obtained from the forward propagation in Section. \ref{sec:forward_prog} is for
\vspace{-0.1cm}
\begin{equation} \label{eq:prior}
\mathbf x_k \boxminus \widehat{\mathbf x}_k = \left( \widehat{\mathbf x}_{k}^\kappa \boxplus \widetilde{\mathbf x}_{k}^\kappa \right) \boxminus \widehat{\mathbf x}_k = \widehat{\mathbf x}_{k}^\kappa \boxminus \widehat{\mathbf x}_{k} + \mathbf J^\kappa \widetilde{\mathbf x}_{k}^\kappa
\end{equation}
where {$\mathbf J^\kappa$} is the partial differentiation of $\left( \widehat{\mathbf x}_{k}^\kappa \boxplus \widetilde{\mathbf x}_{k}^\kappa \right) \boxminus \widehat{\mathbf x}_k$ with respect to $\widetilde{\mathbf x}_{k}^\kappa$ evaluated at zero: 
\vspace{-0.1cm}
\begin{equation}
\begin{split}~\label{e:L_compute}
    \mathbf J^\kappa \!=\!\begin{bmatrix} \mathbf A\!\left(^G \widehat{\mathbf R}_{I_k}^\kappa \!\boxminus\! {}^G \widehat{\mathbf R}_{I_k}\!\right)^{-T} \!\!& \mathbf 0_{3\times 15} \\
\mathbf 0_{15\times 3} & \mathbf I_{15\times 15} \end{bmatrix}
\end{split}
\end{equation}
{where $\mathbf A(\cdot)^{-1}$ is defined in (\ref{e:F_def_parts}).} For the first iteration (i.e., the case of extended Kalman filter), $\widehat{\mathbf x}_{k}^\kappa \!=\! \widehat{\mathbf x}_{k}$, then $\mathbf J^\kappa \!=\! \mathbf I$.

Combining the prior in (\ref{eq:prior}) with the posteriori distribution from (\ref{e:meas_jocob}) yields the maximum a-posteriori estimate (MAP):
\begin{equation}
\begin{split}~\label{e:error_states_solution}
    \min_{\widetilde{\mathbf x}_{k}^\kappa} \left( \| \mathbf x_k \boxminus \widehat{\mathbf x}_k \|^2_{ \widehat{\mathbf P}^{-1}_k} + \sum\nolimits_{j=1}^{m} \| \mathbf z_j^\kappa + \mathbf H_j^\kappa \widetilde{\mathbf x}_{k}^\kappa \|^2_{\mathbf R^{-1}_j} \right)
\end{split}
\end{equation}
where $\| \mathbf x \|_{\mathbf M}^2 = \mathbf x^T \mathbf M \mathbf x$. Substituting the linearization of the prior in (\ref{eq:prior}) into (\ref{e:error_states_solution}) and optimizing the resultant quadratic cost leads to the standard iterated Kalman filter~\cite{qin2019lins}, which can be computed below (to simplify the notation, {let $\mathbf H\! = \![ \mathbf H_1^{\kappa^T}, \cdots, \mathbf H_m^{\kappa^T}]^T$}, $\mathbf R\! =\! \text{diag}\left(\mathbf R_1, \cdots \mathbf R_m \right)$,$\mathbf P\! =\!\left(\mathbf J^{\kappa}\right)^{-1} \widehat{\mathbf P}_{k} (\mathbf J^{\kappa})^{-T} $, and $\mathbf z_k^\kappa = \left[ \mathbf z_1^{\kappa^T}, \cdots, \mathbf z_m^{\kappa^T} \right]^T$):
\vspace{-0.1cm}
\begin{equation}
\begin{split}~\label{e:kalman_gain}
    \mathbf K &= {\mathbf P} \mathbf H^T(\mathbf H {\mathbf P} \mathbf H^T \!+\! \mathbf R)^{-1}, \\
   \widehat{\mathbf x}_{k}^{\kappa+1} &\! \!  = \!  \widehat{\mathbf x}_{k}^{\kappa} \! \boxplus \!  \left( -\mathbf K  {\mathbf z}_k^\kappa  - (\mathbf I - \mathbf K \mathbf H ) (\mathbf J^\kappa)^{-1} \left( \widehat{\mathbf x}_{k}^{\kappa} \boxminus \widehat{\mathbf x}_{k} \right)  \right).
    \end{split}
\end{equation}
The updated estimate $\widehat{\mathbf x}_{k}^{\kappa+1}$ is then used to compute the residual in Section. \ref{sec:residual} and repeat the process until convergence {(i.e., $\| \widehat{\mathbf x}_{k}^{\kappa+1} \boxminus \widehat{\mathbf x}_{k}^\kappa\|\!<\! \epsilon$)}. {After convergence, the optimal state estimation and covariance} is:
\vspace{-0.1cm}
\begin{equation}
\begin{split}~\label{e:state_update}
\bar{\mathbf x}_{k} &= \widehat{\mathbf x}_{k}^{\kappa+1},\ \bar{\mathbf P}_k = \left( \mathbf I - \mathbf K \mathbf H \right) {\mathbf P}
\end{split}
\end{equation}

A problem with the commonly used Kalman gain form in (\ref{e:kalman_gain}) is that it requires to invert the matrix $\mathbf H {\mathbf P} \mathbf H^T \!+\! \mathbf R$ which is in the dimension of the measurements. In practice, the number of LiDAR feature points are very large in number, inverting a matrix of this size is prohibitive. As such, existing works~\cite{qin2019lins, bloesch2017iterated} only use a small number of measurements. In this paper, we show that this limitation can be avoided. The intuition originates from (\ref{e:error_states_solution}) where the cost function is over the state, hence the solution should be calculated with complexity depending on the state dimension. In fact, if directly solving (\ref{e:error_states_solution}), we can obtain the same solution in (\ref{e:kalman_gain}) but with a new form of Kalman gain shown below:
\vspace{-0.1cm}
\begin{equation}
\begin{split}~\label{e:kalman_gain_new}
    \mathbf K \!\!=\!\! \left(\mathbf H^T {\mathbf R}^{-1} \mathbf H \!+\! {\mathbf P}^{-1} \right)^{-1}\!\mathbf H^T \mathbf R^{-1}.
    \end{split}
\end{equation}

We prove in Appendix~\ref{Appendix_Kalman_gain} that the two forms of Kalman gains are indeed equivalent based on the matrix inverse lemma \cite{higham2002accuracy}. Since the LiDAR measurements are independent, the covariance matrix $\mathbf R$ is (block) diagonal and hence the new formula only requires to invert two matrices both in the dimension of state instead of measurements. The new formula greatly saves the computation as the state dimension is usually much lower than measurements in LIO (e.g., {more than 1,000 effective feature points in a scan for 10 $Hz$ scan rate while the state dimension is only 18}). 

\subsubsection{The algorithm}
\
\par
{Our state estimation is summarized in Algorithm \ref{alg:iekf}}.
\vspace{-0.25cm}
\begin{algorithm}
    \caption{{State Estimation}}
    \label{alg:iekf}
    \SetKwInOut{Input}{Input}\SetKwInOut{Output}{Output}\SetKwInOut{Start}{Start}\SetKwInOut{blank}{}
    \Input{Last optimal estimation $\bar{\mathbf x}_{k-1}$ and $\bar{\mathbf P}_{k-1}$, \\
    IMU inputs ($\mathbf a_{m}$, $\bm \omega_m$) in current scan;\\
    LiDAR feature points ${}^{L_j}{\mathbf p}_{f_j}$ in current scan.}
    Forward propagation to obtain state prediction $\widehat{\mathbf x}_k$ via (\ref{e:state_prop}) and covariance prediction $\widehat{\mathbf P}_k$ via (\ref{e:noise_cov});\\
    Backward propagation to obtain ${}^{L_k}{\mathbf p}_{f_j}$ via (\ref{e:back_prop}), (\ref{e:projection});\\
    $\kappa = -1$, $\widehat{\mathbf x}_{k}^{\kappa = 0} = \widehat{\mathbf x}_{k}$; \\
        \Repeat{$\| \widehat{\mathbf x}_{k}^{\kappa+1} \boxminus \widehat{\mathbf x}_{k}^\kappa\|< \epsilon$}{
        $\kappa =\kappa+1$;\\
        Compute $\mathbf J^\kappa$ via (\ref{e:L_compute}) and $\mathbf P\! =\! (\mathbf J^{\kappa})^{-1} \widehat{\mathbf P}_{k} (\mathbf J^{\kappa})^{-T} $;\\
        Compute residual $\mathbf z_j^{\kappa}$ (\ref{e:res_computation}) and Jocobin $\mathbf H_j^\kappa$ (\ref{e:meas_jocob});\\
        Compute the state update $\widehat{\mathbf x}_{k}^{\kappa+1}$ via (\ref{e:kalman_gain}) with the Kalman gain $\mathbf K$ from (\ref{e:kalman_gain_new});\\}
    $\bar{\mathbf x}_{k} = \widehat{\mathbf x}_{k}^{\kappa+1}$; $\bar{\mathbf P}_{k} = \left( \mathbf I - \mathbf K \mathbf H \right) {\mathbf P}$.\\
    \Output{Current optimal estimation $\bar{\mathbf x}_{k}$ and $\bar{\mathbf P}_{k}$.}
\end{algorithm}
            
\subsection{Map Update}
{With the state update $\bar{\mathbf x}_k$ (hence ${}^G\bar{\mathbf T}_{I_k} = ({}^G\bar{\mathbf R}_{I_k}, {}^G\bar{\mathbf p}_{I_k})$), each feature point (${}^{L_k} \mathbf p_{f_j}$) projected to the body frame $L_k$ (see (\ref{e:projection})) is then transformed to the global frame via:}
\begin{equation}
\begin{split}~\label{e:feat_global_map}
    {}^G \bar{\mathbf p}_{f_j} = {}^G\bar{\mathbf T}_{I_k} {}^I\mathbf T_L  {}^{L_k} \mathbf p_{f_j} ; \ j= 1, \cdots, m.
\end{split}
\end{equation}
{These features points are finally appended to the existing map containing feature points from all previous steps.}

\subsection{Initialization}
To obtain a good initial estimate of the system state (e.g., gravity vector ${}^G\mathbf g$, bias, and noise covariance) so to speedup the state estimator, initialization is required. In FAST-LIO, the initialization is simple: keeping the LiDAR static for several seconds {(2 seconds for all the experiments in this paper)}, the collected data is then used to initialize the IMU bias and the gravity vector. If non-repetitive scanning is supported by the LiDAR (e.g., Livox {AVIA}), keeping static also allows the LiDAR to capture an initial high-resolution map that is beneficial for the subsequent navigation.

\section{Experiment Results}\label{sec:experiment}
\subsection{Computational Complexity Experiments}
In order to validate the computational efficiency of the proposed new formula for computing Kalman gains. We intentionally replace the computation of Kalman gains with the old formula in our system and compare their computation time under the same system pipeline and number of feature points. The results are shown in Table.~\ref{tab:c_kalam_runtime}. It is obvious that the complexity of the new formula is much lower than the old one.
\begin{table}[h]
\footnotesize
\centering
\caption{The running times of two Kalman gain formulas}
\label{tab:c_kalam_runtime}
\begin{tabular}{lcccccccc}
\hline
Feature Num. & 307 & 717 & 998 & 1243 & 1453 & 1802 \\ \hline
Old Formula ($ms$) & 7.1 & 23.4 & 109.3 & 251 & 1219 & 1621 \\ \hline
New Formula ($ms$) & 0.07 & 0.11 & 0.25 & 0.37 & 0.59 & 1.16 \\ \hline
\end{tabular}
\end{table}

\subsection{UAV Flight Experiments}
\begin{figure}[t]
    \begin{center}
        {\includegraphics[width=1.0\columnwidth]{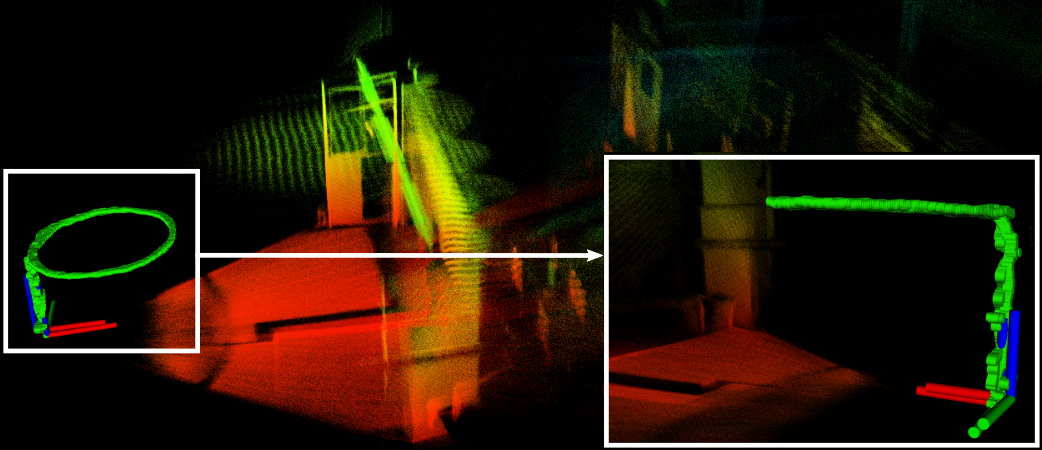}}
    \end{center}
    \vspace{-0.25cm}
    \caption{\label{fig:uav_indoor_50Hz}{During the flight experiment, the UAV is automatically flying in a circle path with 1.8 $m$ radius and 1.4 $m$ height. The circle path is conducted repeatedly for 4 times with different periods (6-10 $s$). The yaw command of the UAV maintains constant during the flight. In the end, the UAV is manually controlled to land at the take-off point, which enables us to measure the drift. }}
\end{figure}

In order to validate the robustness and computational efficiency of FAST-LIO in actual mobile robots, we build a small-scale quadrotor that can carry a Livox Avia LiDAR with 70$^\circ$ FoV and a DJI Manifold 2-C onboard computer with a {1.8 $GHz$ quad-core Intel i7-8550U CPU} and {8 $GB$ RAM}, as shown in Fig.~\ref{fig:uav_fly}. The UAV has only 280 $mm$ wheelbase, and the LiDAR is directly installed on the airframe. The LiDAR-inertial odometry is sent to the flight controller tracking a circle trajectory (Fig. \ref{fig:uav_indoor_50Hz}). The actual flight experiments show that FAST-LIO can achieve real-time and stable odometry output and mapping {in a maximum of 50 $Hz$ for the indoor environment. The flight trajectory and mapping results of 50 $Hz$ frame rate indoor experiment are shown in Fig.~\ref{fig:uav_indoor_50Hz}. The average number of effective feature points and running time is 270 and 6.7 $ms$, respectively , the drift is smaller than 0.3\% (0.08 $m$ drift over 32 $m$ trajectory). The flight video can be found at} \url{https://youtu.be/iYCY6T79oNU}.
\subsection{Indoor Experiments}\label{sec:indoor_test}
\begin{figure}[t]
    \begin{center}
        {\includegraphics[width=1.0\columnwidth]{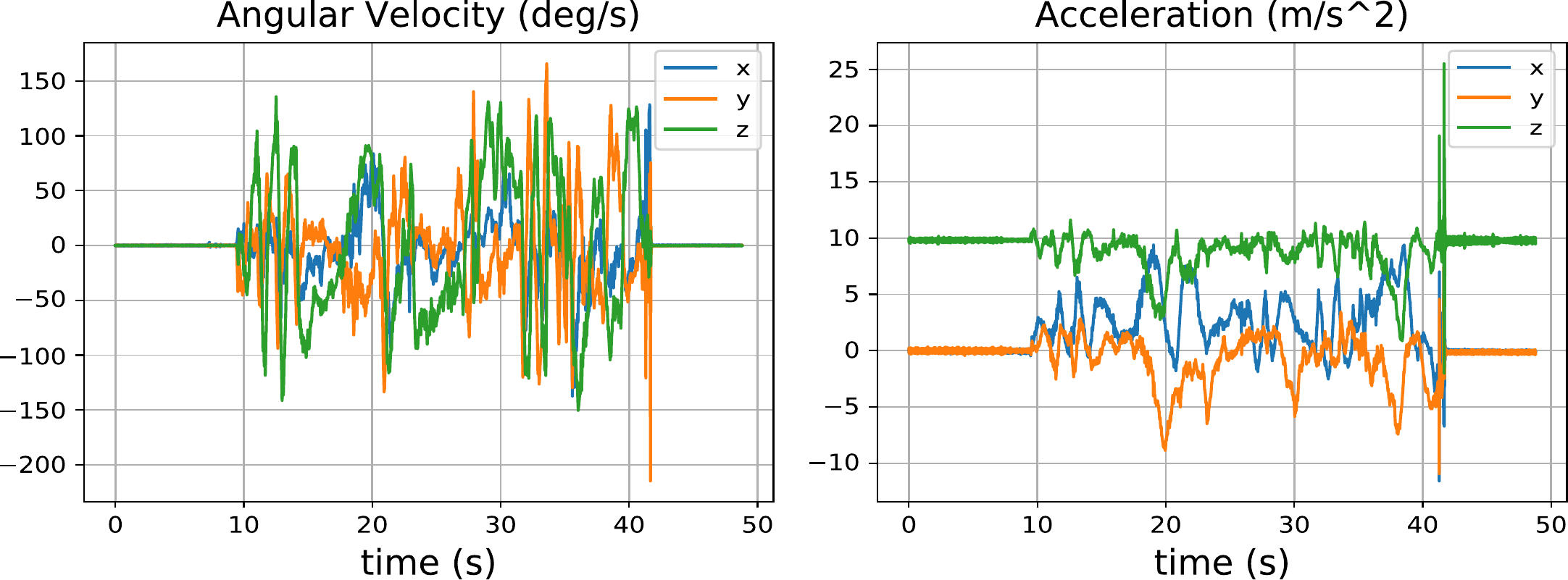}\vspace{-0.3cm}}
    \end{center}
    \caption{\label{fig:indoor_imu}The angular velocity and acceleration in the indoor experiments.}
\end{figure}

\begin{figure}[t]
    \begin{center}
        {\includegraphics[width=1.0\columnwidth]{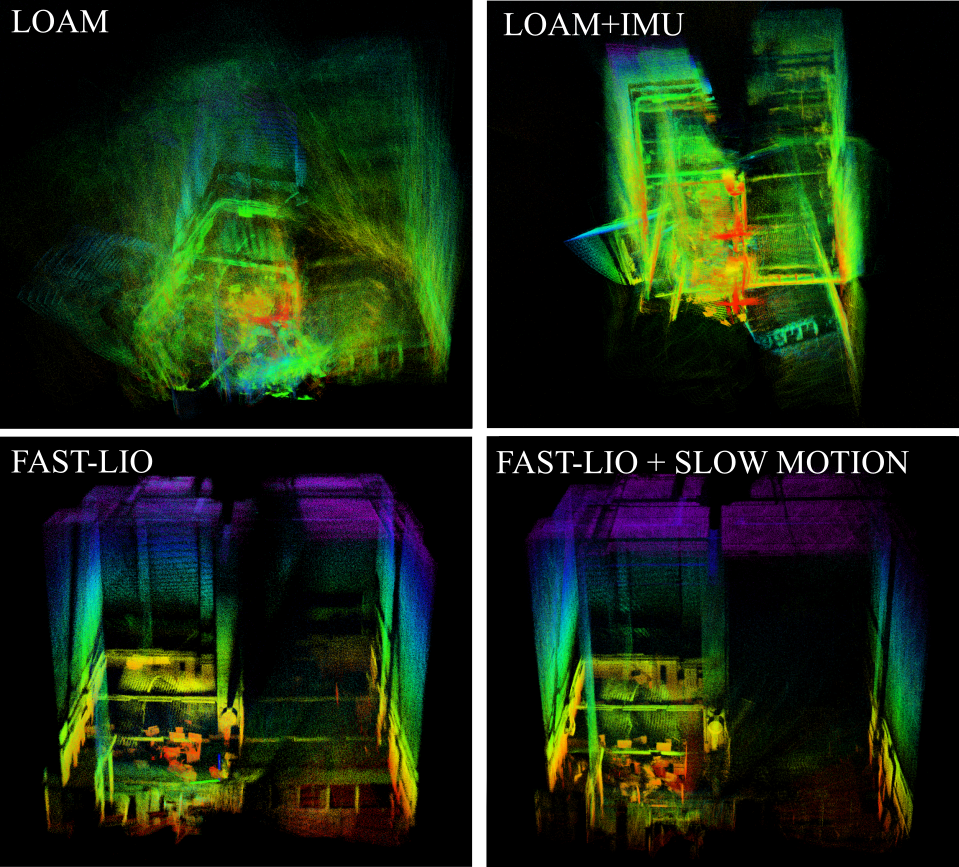}\vspace{-0.25cm}}
    \end{center}
    \caption{\label{fig:indoor_map}The Mapping results of different LIO packages in an indoor environment with large rotation speed.}
\end{figure}

\begin{table}[t]
	\footnotesize
	\centering
	\caption{Comparison of processing time for a LiDAR scan at $10Hz$}\label{tab:c_runtime}
	\begin{tabular}{ccccc}
		\hline
		Packages & Num. of effective features & Running time \\ \hline
		LOAM     & 1107 & 59 $ms$ \\  \hline
		LOAM+IMU & 1107 & 44 $ms$ \\ \hline
		FAST-LIO & 1430 & 23 $ms$\\ \hline
	\end{tabular}
	\vspace{-0.cm}
\end{table}
Then we test FAST-LIO in a challenging indoor environment with large rotation speeds. In order to generate large rotation, the sensor suite is held on hands and undergoes quickly shaking. Fig.~\ref{fig:indoor_imu} shows the angular velocity and acceleration during the experiment. It is seen that the angular velocity often exceeds $100\ deg/s$. A state of the art implementation of LOAM on Livox LiDARs\footnote[5]{\url{https://github.com/Livox-SDK/livox_mapping}}\cite{lin2019loam_livox} {and} LOAM with IMU\footnote[6]{\url{https://github.com/Livox-SDK/livox_horizon_loam}}\cite{zhang2014loam} are also tested as comparisons when the feature extraction are replaced with the one of FAST-LIO. The results show that FAST-LIO can output odometry faster and more stable than others, as shown in Fig.~\ref{fig:indoor_map} and Table.~\ref{tab:c_runtime}. It should be noted that the LOAM+IMU is a loosely-coupled method, hence results in inconsistent mapping. To further verify the mapping result, we perform a second experiment in the same environment but with a much slower motion. The map built by FAST-LIO is shown in the lower-right figure of Fig.~\ref{fig:indoor_imu}.  Since the two experiments have non-identical movements, it leads to slight visual differences at places occlusions occur. The rest mapping results are very close.

\subsection{Outdoor Experiments}
\begin{figure}[t]
    \begin{center}
        {\includegraphics[width=1\columnwidth]{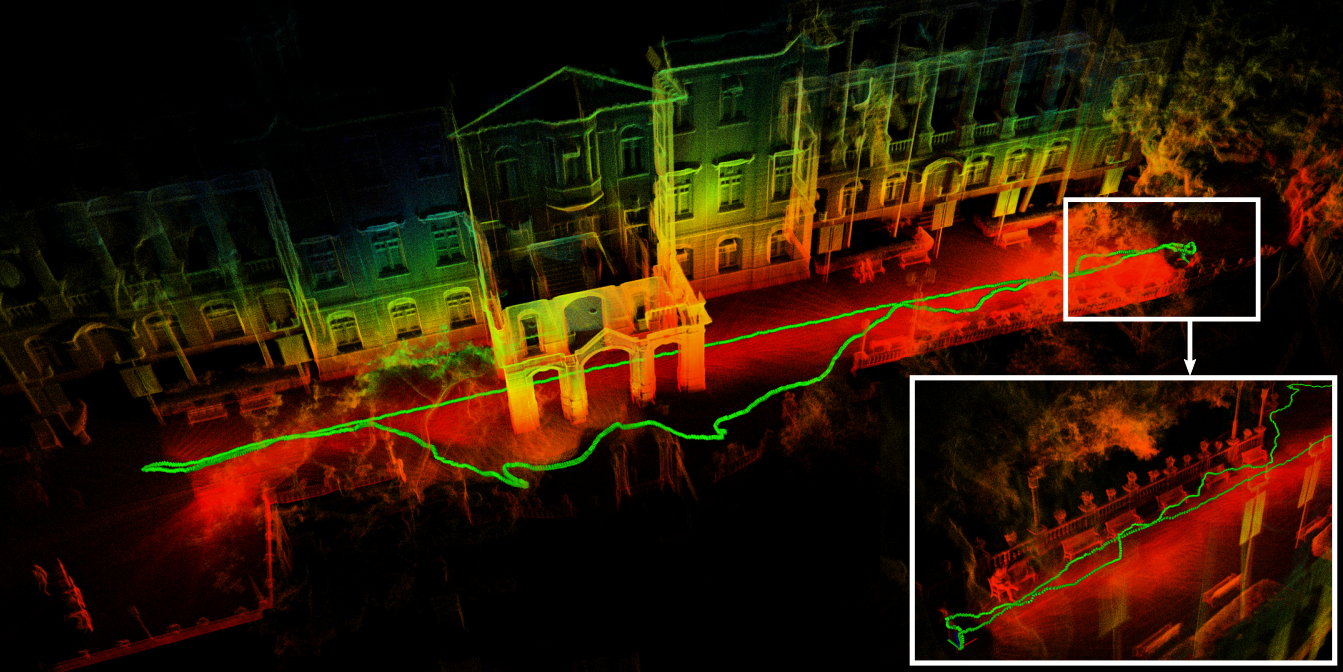}\vspace{-0.3cm}}
    \end{center}
    \caption{\label{fig:outdoor_hku_mb}Mapping results of the Main Building, University of Hong Kong. The LiDAR platform, the same one in Fig.~\ref{fig:uav_fly}, is handheld randomly walking to scan the building. In order to show the drift, the experiment are started and ended at the same place.}
    \vspace{-0.25cm}
\end{figure}

Here we show the performance of FAST-LIO in outdoor environments. Fig.~\ref{fig:outdoor_hku_mb} shows the mapping results (displaying all raw points) of the Main Building in the University of Hong Kong. The sensor suite is handheld during the data collection and returned to the starting position after traveling around $140m$. {The drift in this experiment is smaller than 0.05\% (0.07 $m$ drift over 140 $m$ trajectory).} The scan rate is set to 10 $Hz$ in this experiment, and the average processing time of a scan is {25 $ms$} with average 1497 effective feature points.

\begin{figure}[t]
    \begin{center}
        {\includegraphics[width=1\columnwidth]{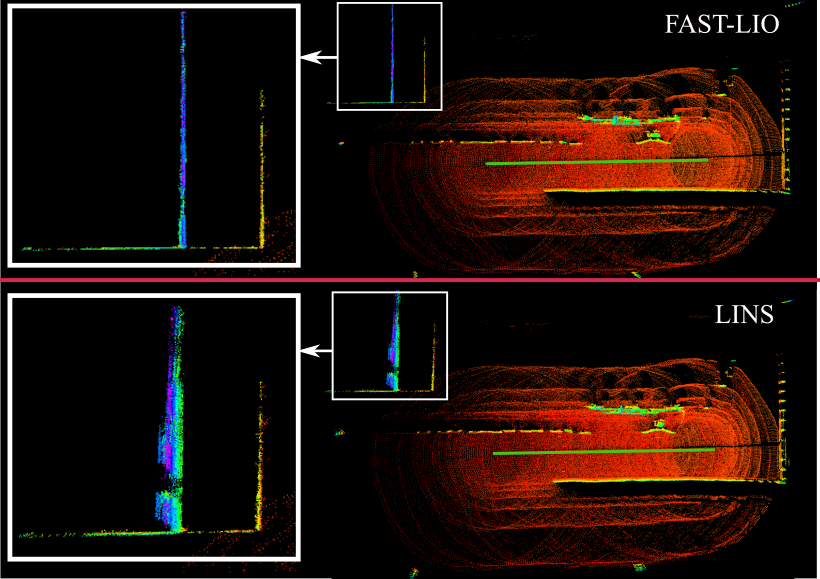}\vspace{-0.5cm}}
    \end{center}
    \caption{\label{fig:lins} Comparison between LINS \cite{qin2019lins} and FAST-LIO. The data is from \cite{qin2019lins} and is collected by a Velodyne VLP-16 LiDAR and an Xsens MTiG-710 IMU, the green straight line in the center is the odometry output.}
    \vspace{-0.5cm}
\end{figure}

{Further, we compare FAST-LIO with LINS}\footnote[7]{\url{https://github.com/ChaoqinRobotics/LINS---LiDAR-inertial-SLAM}}\cite{qin2019lins}. {To make a fair comparison, we use the dataset from LINS\cite{qin2019lins}, which is a seaport area data collected by a Velodyne VLP-16 and an Xsens MTiG-710 IMU$^7$. The results show that the FAST-LIO can achieve better mapping accuracy (see Fig.~\ref{fig:lins}) and only consumes 7.3 $ms$ processing time in average while LINS takes 34.5 $ms$ in average, both running at $10Hz$. It should be noted that since the EKF formula in LINS package has high computational complexity (see Section.~\ref{sec:iter-update})), it down samples the feature points to average 147 points in a scan (while 784 in a scan for FAST-LIO). This leads to degraded mapping accuracy for LINS. The result in Fig.~\ref{fig:lins} shows all the feature points (before down-sample) of FAST-LIO and LINS. All the experiments are conducted on the DJI Manifold2 onboard computer.}

\section{Conclusion}\label{sec:conclusion}
This paper proposed FAST-LIO, a computationally efficient and robust LiDAR-inertial odometry framework by a tightly-coupled iterated Kalman filter. We used the forward and backward propagation to predict the states and compensate for the motion in a LiDAR scan. Besides, we proved and implemented an equivalent formula that can achieve much lower complexity for the Kalman gain computation. FAST-LIO was tested in the UAV flight experiment, challenging indoor environment with large rotation speed and outdoor environment. In all tests, our method produced precise, real-time, and reliable navigation results.
\appendix
\subsection{Computation of $\mathbf F_{\widetilde{\mathbf x}}$ and $\mathbf F_{\mathbf w}$} \label{Appendix_F_comp}
Recall  ${\mathbf x}_{i} = \widehat{\mathbf x}_{i} \boxplus \widetilde{\mathbf x}_i$, denote $\mathbf g\left(\widetilde{\mathbf x}_i, \mathbf w_i\right)=\mathbf f({\mathbf x}_i,\mathbf u_i, \mathbf w_i)\Delta t=\mathbf f(\widehat{\mathbf x}_i\boxplus\widetilde{\mathbf x}_i,\mathbf u_i, \mathbf w_i)\Delta t$. Then the error state model (\ref{e:error_state_model}) is rewriten as:
\begin{equation}
\begin{split}~\label{e:delta_x_def}
\widetilde{\mathbf x}_{i+1} = \underbrace{\left(\left( \widehat{\mathbf x}_{i}\! \boxplus\! \widetilde{\mathbf x}_i\right)\! \boxplus\!\mathbf g\left( \widetilde{\mathbf x}_{i}, \mathbf w_i\right)\right) \boxminus \left(\widehat{\mathbf x}_{i}\! \boxplus\! \mathbf g\left( \mathbf 0, \mathbf 0 \right) \right)}_{\mathbf G\left(\widetilde{\mathbf x}_i, \mathbf g\left(\widetilde{\mathbf x}_i, \mathbf w_i\right) \right)}
\end{split}
\end{equation}
Following the chain rule of partial differention, the matrix $\mathbf F_{\widetilde{\mathbf x}}$ and $\mathbf F_{\mathbf w}$ in (\ref{e:error_state_model}) are computed as below.
\begin{equation}
\begin{split}~\label{e:F_def}
\mathbf F_{\widetilde{\mathbf x}} &=\left. \textstyle \left(\frac{\partial \mathbf G\left(\widetilde{\mathbf x}_i, \mathbf g\left(\mathbf 0, \mathbf 0\right) \right)}{\partial \widetilde{\mathbf x}_{i}}+\frac{\partial \mathbf G\left(\mathbf 0, \mathbf g\left(\widetilde{\mathbf x}_i, \mathbf 0 \right) \right)}{\partial \mathbf g\left(\widetilde{\mathbf x}_i, \mathbf 0 \right)}\frac{\partial \mathbf g\left(\widetilde{\mathbf x}_i, \mathbf 0 \right)}{\partial \widetilde{\mathbf x}_i}\right)\right|_{\begin{subarray}{l}\widetilde{\mathbf x}_i = \mathbf 0\end{subarray}} \\
\mathbf F_{\mathbf w} &=\left. \textstyle \left(\frac{\partial \mathbf G\left(\mathbf 0, \mathbf g\left(\mathbf 0, \mathbf w_i\right) \right)}{\partial \mathbf g\left(\mathbf 0, \mathbf w_i\right)}\frac{\partial \mathbf g\left(\mathbf 0, \mathbf w_i\right)}{\partial \mathbf w_i}\right)\right|_{\begin{subarray}{l}\mathbf w_i = \mathbf 0\end{subarray}}
\end{split}
\end{equation}

\subsection{Equivalent Kalman Gain formula} \label{Appendix_Kalman_gain}
Based on the matrix inverse lemma~\cite{higham2002accuracy}, we can get:
\begin{equation}
\begin{split}~\label{e:inverse_lemma}\notag
\left(\mathbf P^{-1}\!+\!\mathbf H^T {\mathbf R}^{-1} \mathbf H \right)^{-1} \!=\! \mathbf P - \mathbf P \mathbf H^T \left( \mathbf H \mathbf P \mathbf H^T + \mathbf R \right)^{-1} \mathbf H \mathbf P
\end{split}
\end{equation}

Substituting above into (\ref{e:kalman_gain_new}), we can get:
\begin{equation}
\begin{split}~\label{e:kalman_gain_new_proof}\notag
    \mathbf K \!&=\! \left(\mathbf H^T {\mathbf R}^{-1} \mathbf H \!+\! {\mathbf P}^{-1} \right)^{-1}\!\mathbf H^T \mathbf R^{-1} \\
    &=\!\mathbf P \mathbf H^T \mathbf R^{-1} \!-\! \mathbf P \mathbf H^T \left( \mathbf H \mathbf P \mathbf H^T \!+\! \mathbf R \right)^{-1} \mathbf H \mathbf P\!\mathbf H^T \mathbf R^{-1}
    \end{split}
\end{equation}

Now note that $\mathbf H\mathbf P\mathbf H^T\mathbf R^{-1} \!=\! \left(\mathbf H\mathbf P\mathbf H^T + \mathbf R \right)\mathbf R^{-1} \!-\! \mathbf I$. Substituting it into above, we can get the standard Kalman gain formula in (\ref{e:kalman_gain}), as shown below.
\begin{equation}
\begin{split}~\label{e:kalman_gain_final} \notag
    \mathbf K &=\mathbf P \mathbf H^T \mathbf R^{-1} - \mathbf P \mathbf H^T \mathbf R^{-1} + \mathbf P \mathbf H^T \left( \mathbf H \mathbf P \mathbf H^T + \mathbf R \right)^{-1} \\
    &= \mathbf P \mathbf H^T \left( \mathbf H \mathbf P \mathbf H^T + \mathbf R \right)^{-1}. \quad \blacksquare
    \end{split}
\end{equation}

\bibliography{paper}
\end{document}